\documentclass[lettersize,journal]{IEEEtran}
\usepackage{amsfonts}
\usepackage[T1]{fontenc}
\usepackage{aecompl}
\usepackage{array}
\usepackage{textcomp}
\usepackage{stfloats}
\usepackage{url}
\usepackage{verbatim}
\usepackage{graphicx}
\usepackage{cite}
\usepackage{float}
\usepackage{mathtools}
\usepackage{balance}
\usepackage{booktabs}
\usepackage{subfig}
\usepackage{caption}
\usepackage{amsmath, amssymb}
\usepackage{booktabs}
\usepackage{pifont}
\usepackage[section]{placeins}
\usepackage{xcolor}
\usepackage{algorithmic}
\usepackage{algorithm}
\definecolor{ADAPG}{rgb}{1,0.7019607843137255,0.7019607843137255}
\definecolor{Adamw}{rgb}{0.0, 0.9, 0.0}

\begin{document}

\title{}
\title{Adaptive Diffusion Policy Optimization for Robotic Manipulation}
% \title{Effective Reinforcement Learning with Adaptive Policy Gradient}

\author{Huiyun Jiang, Zhuang Yang~\IEEEmembership{Member,~IEEE}
        % <-this % stops a space
\thanks{(Corresponding authors: Zhuang Yang). Huiyun Jiang and Zhuang Yang are with the School of Computer Science and Technology, Soochow University, Suzhou 215006, China. (e-mail: HuiyunJiang@163.com; zhuangyng@163.com)}}

\markboth{Journal of \LaTeX\ Class Files,~Vol.~14, No.~8, August~2021}%
{Shell \MakeLowercase{\textit{et al.}}: A Sample Article Using IEEEtran.cls for IEEE Journals}


\maketitle
\pagestyle{empty}  % no page number for the second and the later pages
\thispagestyle{empty} % no page number for the first page
\begin{abstract}

Recent studies have shown the great potential of diffusion models in improving reinforcement learning (RL) by modeling complex policies, expressing a high degree of multimodality, and efficiently handling high-dimensional continuous control tasks. However, there is currently limited research on how to optimize diffusion-based polices (e.g., Diffusion Policy) fast and
stably. In this paper, we propose an Adam-based Diffusion Policy Optimization (ADPO), a fast algorithmic framework containing best practices for fine-tuning diffusion-based polices in robotic
control tasks using the adaptive gradient descent method in RL. Adaptive gradient method is less studied in training RL, let alone diffusion-based policies. We confirm that ADPO outperforms
other diffusion-based RL methods in terms of overall effectiveness for fine-tuning on standard robotic tasks. Concretely, we conduct extensive experiments on standard robotic control tasks
to test ADPO, where, particularly, six popular diffusion-based RL methods are provided as benchmark methods. Experimental results show that ADPO acquires better or comparable performance than the baseline methods. Finally, we systematically analyze the sensitivity of multiple hyperparameters in standard robotics tasks, providing guidance for subsequent practical applications. Our video demonstrations are released in https://github.com/Timeless-lab/ADPO.git.
\end{abstract}

\begin{IEEEkeywords}
Diffusion policy, Reinforcement learning, Policy gradient, Robotic control.
\end{IEEEkeywords}

\section{Introduction}
\IEEEPARstart{R}{einforcement} learning (RL) is a machine learning approach that enables agents to learn and make decisions by interacting with its environment. As artificial intelligence (AI) drives a new wave of scientific and technological advancements, RL has emerged as a key method for robot operation and has achieved remarkable success in the field of robot control. For example, Amazon’s DeepRacer platform\cite{petryshyn2024deep} uses RL to train autonomous driving models; the AdaRL-MDF framework \cite{qi2023adaptive} uses adaptive reinforcement learning to train a robot to play the Rock–Paper–Scissors (RPS) game with humans; ABC\_RL \cite{cui2022reinforcement}, an artificial bee colony (ABC) algorithm based on RL, significantly enhances the efficiency of robots in path planning. These achievements highlight RL's vast potential across diverse applications.

Although RL has made significant advancements in robotics, it still encounters several limitations and challenges.
For instance, in terms of policy parameterization, the use of Gaussian distributions \cite{sutton1999policy} or Gaussian Mixture Models (GMM) \cite{bishop2006pattern} in traditional RL methods tend to generate unimodal distributions in the action space, limiting the agent’s ability to explore the environment and weakening the expression of complex policies. These limitations are particularly evident in robotic control tasks, where the state and action spaces are complex and continuous, often resulting in unstable policy updates.

As a powerful generative model, the diffusion model\cite{ho2020denoising,chen2023diffusiondet,xu2023versatile} excels at learning complex distributions and has demonstrated outstanding performance in behavior cloning, showing the much promise in addressing the limitations and challenges of RL. To name a few, Chi et al. \cite{chi2023diffusion} introduce a novel method for behavior policy generation by using the conditional denoising process of the diffusion model to represent robot behavior policies. Compared to traditional approaches in RL, this method more effectively handles multimodal action distributions while offering superior stability and expressiveness. 

Surprisingly, we find that current research on optimizing diffusion policy remains relatively limited. At present, Ren et al. \cite{ren2024diffusion} develop Diffusion Policy Policy Optimization (DPPO) leverages the policy gradient (PG) method to fine-tuning the diffusion policy and use proximal policy optimization (PPO) to improve the PG. Gao et al. \cite{gao2025behavior} propose behavioral regularized diffusion policy optimization (BDPO) to improve the diffusion policy by extending behavioral Kullback-Leibler (KL) regularization in the diffusion generation path. Ding et al. \cite{ding2024diffusion} propose Q-weighted variational policy optimization (QVPO), where specifically, QVPO uses the Q-weighted VLO loss as a strict lower bound for optimizing the diffusion policy. We emphasize that the methods mentioned above all optimize the policy update calculation formula, either directly or indirectly.

Motivated by the recent development of diffusion model with RL, this work proposes a diffusion-based enhanced RL framework and introduce an adaptive policy gradient mechanism, which is totally different from the methods mentioned above. Concretely, this mechanism introduces a discount factor into adaptive gradient methods, thus making the resulting algorithms interpolate between different optimizers (such as Adam \cite{kingma2014adam}, RMSProp \cite{tieleman2012lecture}, etc.). Such operation further improves the policy optimization performance while retains their respective advantages.

For clarity and ease to be comprehended, we summarize the main \textbf{contributions} of this paper as follows:

\begin{itemize}
\item We introduce the Adam-based Diffusion Policy Optimization (ADPO) framework, designed to accelerate diffusion model-based RL methods. To the best of our knowledge, this is the first proposal to leverage Adaptive Policy Gradient (ADAPG) to enhance the performance of diffusion-based RL.
\item We compare ADPO with several existing diffusion-based RL methods, including but not limited to Diffusion Policy Policy Optimization (DPPO) \cite{ren2024diffusion}, Diffusion Advantage-Weighted Regression (DAWR) \cite{peng2019advantage}, Model-free online RL with Diffusion Policy (DIPO) \cite{yang2023policy}, on standard robotic tasks. Numerical experiments on all standard robotic tasks demonstrate that ADPO significantly outperforms in terms of both training stability and final policy performance.
\end{itemize}

\section{Related Work}
This section concisely reviews the recent development of Diffusion-based RL and ADAPG. 
\subsection{Diffusion-based RL}
\label{DMRL}
Recently, the application of diffusion models in RL has become increasingly popular and has shown great potential in policy expression.
In offline RL, we often face the problem of being unable to fit the distribution of data sets or a lack of diversity. To address this problem, subsequent works \cite{chen2022offline,wang2022diffusion,hansen2023idql,ada2024diffusion,kang2023efficient} parameterize the policy using diffusion model to capture multimodal distributions, thereby enhancing the exploration ability and generalization of the policy and alleviating the error between the cloned behavior policy and the true behavior policy. Ajay et al. \cite{ajay2022conditional} formulate sequential decision-making as a conditional generative modeling problem and use a classifier-free guided low-temperature sampling to obtain the maximum reward. Li et al. \cite{li2023hierarchical} propose a classifier-free guided hierarchical trajectory diffusion model (HDMI), which uses a reward-conditioned model to discover sub-goals and a goal-conditioned diffuser to generate action sequences. Zhu et al. \cite{zhu2024madiff} propose using an attention-based diffusion model to fit complex movements between multiple agents and achieve information interaction between agents. Recently, studies have also shown that diffusion models can promote online RL training. Wang et al. \cite{wang2024diffusion} propose a diffusion factor criticism algorithm with an entropy regulator, which enhances the representation ability of the policy by estimating the diffusion policy using a Gaussian mixture model. 

\subsection{ADAptive Policy Gradient}
\label{ADAPG}
Currently, policy optimization methods in reinforcement learning (e.g., soft actor-critic (SAC) \cite{haarnoja2018soft}, proximal policy optimization (PPO) \cite{schulman2017proximal}, and trust region policy optimization (TRPO) \cite{schulman2015trust}) have made significant progress. These methods are all aimed at directly optimizing the agent's policy so that it can choose more optimal actions in the environment and achieve higher long-term rewards.
However, despite the strong performance of existing policy optimization methods in many tasks, the training process may still encounter issues such as slow convergence or instability in certain cases. To address these challenges and accelerate the path to optimal solutions, the ADAPG method is introduced. ADAPG utilizes the adaptive gradient method in machine learning and adjusts the weighted average of different algorithms (such as RMSProp \cite{tieleman2012lecture} and Adam \cite{kingma2014adam}) by introducing an immediate discount factor \(\lambda\) to achieve better performance. At the same time, to address the error accumulation caused by noise, ADAPG considers the Katyusha \cite{allen2018katyusha} momentum scheme to reduce oscillations in the parameter update process and avoid converging to local optimality.

\section{Preliminaries}
Some fundamental information about RL and diffusion model are provided in this section.
\subsection{Reinforcement Learning}
\label{RL}
In order to accomplish a task, RL\cite{thrun2000reinforcement} addresses sequential decision-making problems by training an agent to maximize cumulative rewards. Generally, the environment in RL is modeled as a Markov Decision Process (MDP), recorded briefly  \( M = \{ S, A, P, R, \gamma, O \} \)with state space \(S\), action space \(A\), 
environment dynamics \( P(\mathbf{s'} \mid \mathbf{s}, \mathbf{a}) \), reward function \( R : S \times A \rightarrow \mathbb{R} \), discount factor \(\gamma \in [0,1)\), and initial state distribution \(O\).

At each time step \( t \), the agent observes the current state \( s_t \), takes an action \( a_t \) according to the policy \(\pi(a_t\mid s_t)\), and transitions to the next state \( s_{t+1} \) based on the state transition probability \(P(s_{t+1} \mid s_t, a_t)\), at the same time. Finally, the agent receives a reward \(R(s_t, a_t)\). The goal of the agent  is to learn policy \( \pi_{\theta}(a \mid s) \), parameterized by \( \theta \), through maximizing the following cumulative discounted reward:
\begin{equation}
\label{6}
G_{\pi_{\theta}} = \mathbb{E} \left[ \sum_{t=0}^{\infty} \gamma^t r(s_t, a_t) \right].
\end{equation}

Under the policy \(\pi_{\theta}\), the value function is used to measure the long-term benefit of a state \(s\) (referred to as the state-value function) or a state-action pair \((s,a)\) (referred to as the action-value function). Typically, the optimal policy is solved by maximizing the value function(\(V_{\pi_{\theta}}^*(s)\) or \(Q_{\pi_{\theta}}^*(s,a)\)) and deriving it using the Bellman equation:
\begin{equation}
\label{7}
V_{\pi_{\theta}}^*(s) = \max_{a\in \mathcal{A}} \sum_{s' \in \mathcal{S}} p(s' \mid s, a) \left[ r(s, a) + \gamma V_{\pi_{\theta}}^*(s')\right],
\end{equation}
\begin{equation}
\label{8}
Q_{\pi_{\theta}}^*(s,a) = \sum_{s' \in \mathcal{S}} p(s' \mid s, a) \left[ r(s, a) + \gamma \max_{a'\in \mathcal{A}}Q_{\pi_{\theta}}^*(s',a')\right].
\end{equation}

The value function introduced above is a value-based method, which seeks the optimal policy by maximizing value. Another method is a policy-based method (e.g., Eq.~\eqref{9}), which directly optimizes the policy, \( \pi_{\theta}(a \mid s) \), to maximize the cumulative reward,
\begin{equation}
\label{9}
\nabla_{\theta} J(\theta) = \mathbb{E} \left[ \sum_{t=0}^{\infty} \nabla_{\theta} \log \pi_{\theta} (a_t \mid s_t) R(a_t, s_t) \right],
\end{equation}
Notice that, to reduce variance and improve stability of policy gradient methods, usually using the advantage function \(A_{\pi_{\theta}}(s, a) = R(a_t,s_t) - V_{\pi_{\theta}}(s_t)\) substitutes \(R(a_t, s_t)\).

\subsection{Diffusion Model}
\label{DM}
The diffusion model is a powerful generative model distinguished by its two-phase process: forward noise addition and backward denoising. This unique mechanism sets it apart from other generative models and endows it with strong probabilistic modeling capabilities\cite{cao2024survey,yang2023diffusion,song2020score}, allowing it to accurately learn complex data distributions and generate high-quality samples.

A denoising diffusion probabilistic model (DDPM) \cite{ho2020denoising,sohl2015deep}, the representative one of diffusion model, consists of two parameterized Markov chains, the forward chain and the reverse Markov chain.

The forward chain gradually adds noise to the original data, perturbing it into pure noise that approximates a Gaussian distribution. Specifically, given a data distribution \( x_0 \sim q(x_0) \), the forward Markov chain progressively adds noise, $\epsilon$,  to \( x_0 \) over \( T \) time steps, transforming it into noise \( x_T \) that closely follows a standard Gaussian distribution. 
\begin{equation}
\label{1}
q(\mathbf{x}_t \mid \mathbf{x}_0) = \mathcal{N}(\mathbf{x}_t; \sqrt{\bar{\alpha}_t} \mathbf{x}_0, (1 - \bar{\alpha}_t) \mathbf{I}).
\end{equation}

In contrast, the reverse Markov chain reconstructs the original data by reversing the forward process through the neural network parameterization. The neural network predicts the added noise $\epsilon$ when transforming \( x_t \) to \( x_0 \). The specific formula for the sampling process is as follows:
\begin{equation}
\label{2}
p_{\theta}(\mathbf{x}_{t-1} \mid \mathbf{x}_t) = \mathcal{N}(\mathbf{x}_{t-1}; \mu_{\theta} (\mathbf{x}_t, t), \Sigma_{\theta}(\mathbf{x}_t, t)).
\end{equation}

During the training of the diffusion model, the aim is to minimize the loss function by reducing the discrepancy between the data distributions \(q(x)\) and \(p_{\theta}(x)\) through network optimization:

\begin{equation}
\label{3}
Loss = \mathbb{E} \left[ \left\| {\epsilon}^{t} - {\mu_{\theta}}^{t} (x_t, t) \right\|^2 \right].
\end{equation}

\subsection{Diffusion model as policies}
\label{Diffusion models as policies}
The diffusion policy \cite{ren2024diffusion} is a DDPM that parameterizes the behavior policy \( \pi_\theta \). In order to enable the DDPM to learn the robot's motion, the diffusion policy uses state, \( s \), in RL as a conditional parameter, and the output \( x \) represents the robot's action. Thus, instead of Eq. \eqref{2} in the original diffusion model, the DDPM is now used to approximate the conditional distribution \( p_{\theta}(a_{k-1} \mid a_k, s) \), where, particularly, the modified backward chain is shown in Eq.~\eqref{4}.
\begin{equation}
\label{4}
p_{\theta}(\mathbf{a}_{k}^{t-1} \mid \mathbf{a}_{k}^{t}, \mathbf{s})) = \mathcal{N}(\mathbf{a}_{k}^{t-1}; \mu_{\theta}(\mathbf{a}_{k}^{t},\mathbf{s}, t), \Sigma_{\theta}(\mathbf{a}_{k}^{t},\mathbf{s}, t)).
\end{equation}
In Eq.~\eqref{4}, diffusion policy has two layers of Markov chains, \( t \) represents the Markov chain of the internal denoising process of the diffusion model, and \( k \) represents the Markov chain of the environment in external RL. By embedding the diffusion MDP into the MDP of the environment, a larger diffusion policy MDP is obtained. Therefore, the training loss of Eq.~\eqref{3} turns out to
\begin{equation}
\label{5}
Loss = \mathbb{E} \left[ \left\| {\epsilon}^{t} - {\mu_{\theta}}^{t} ({s_k} ,a_k^0 + {\epsilon}^{t}, t) \right\|^2 \right].
\end{equation}

\section{Method}
In this section, we first introduce the details of six popular Diffusion-based RL methods in subsection \ref{SIX}. Then, we introduce our Adam-based Diffusion Policy Optimization framework in subsection \ref{ADPOFRAMEWORK}.
\subsection{Six popular Diffusion-based RL methods}
\label{SIX}
Diffusion Policy Policy Optimization (DPPO) \cite{ren2024diffusion}: in DPPO, the diffusion policy is enhanced by incorporating the policy gradient method:
\begin{equation}
\label{10}
\begin{split}
\mathcal{L}_{actor} = - \mathbb{E}_{\pi_{\theta}} \left[{\sum_{k=0}^{T} \log \pi_{\theta} (a_k^0 | s_k){A}_{\pi_{\theta}} (s_k, a_k^0) }\right].
\end{split}
\end{equation}

To address instability during policy updates and maintain similarity between the new and old policy distributions, DPPO further incorporates proximal policy optimization (PPO), a widely used algorithm for policy gradient updates:
\begin{equation}
\label{11}
\begin{split}
\mathcal{L}_{actor} = \mathbb{E}_{\pi_{\theta}'} 
\min \Bigg( 
{A}_{\pi_{\theta}'} (s_k, a_k^0) 
\frac{\pi_{\theta} (a_k^0 | s_k)}{\pi_{\theta}' (a_k^0 | s_k)},  \\
{A}_{\pi_{\theta}'} (s_k, a_k^0) 
\text{clip} \left( \frac{\pi_{\theta} (a_k^0 | s_k)}{\pi_{\theta}' (a_k^0 | s_k)}, 1 - \varepsilon, 1 + \varepsilon \right) 
\Bigg).
\end{split}
\end{equation}
Above, \(\pi_{\theta}'\) denotes the old policy, while \(\pi_{\theta}\) represents the pruned new policy. The pruning ratio $\epsilon$ regulates the update magnitude from the old policy to the new one. Additionally, the advantage function in DPPO is defined as follows: 
\(A_{\pi_{\theta}'}(s_k,a_k^0) = \gamma_{\text{denoise}}(R(s_k,a_k^0) - \hat V_{\phi}(s_k))\), 
where \(\gamma_{\text{denoise}} \in (0,1)\) denotes denoising discount.

In the value function, the critic is trained to minimize the discrepancy between the predicted value and the target value:
\begin{equation}
\label{12}
\begin{split}
\mathcal{L}_{critic} = \mathbb{E}_{\phi} \big[ \| \hat{V}_{\phi}(s_k,a_k^0) - V(s_k,a_k^0)\|^2\big].
\end{split}
\end{equation}

Diffusion Advantage-Weighted Regression (DAWR) \cite{peng2019advantage}: AWR is a simple off-policy algorithm for model-free reinforcement learning. It is based on the idea of reward-weighted regression (RWR) \cite{peters2007reinforcement}. Specifically, DAWR updates the diffusion policy using TD-bootstrapped advantage estimates, ensuring that the new policy generates a more optimal action distribution compared to the previous one. DAWR's action policy is updated as follows:
\begin{equation}
\label{14}
\begin{split}
\mathcal{L}_{actor} = - \mathbb{E}_{\pi_{\theta}} \left[{\sum_{k=0}^{T} \log \pi_{\theta} (a_k^0 | s_k)\exp\left(\hat A_{\phi}(s_t,a_t^0) \right)}\right].
\end{split}
\end{equation}

% \textcolor{red}{Note that the} AWR critic network update follows the same formulation as Eq.~\eqref{12}.

Model-free online RL with Diffusion Policy (DIPO) \cite{yang2023policy}: DIPO proposes a new “action gradient” to improve diffusion policy. The data is updated by rewriting the exponential integral discrete diffusion policy and put into the replay buffer \(D\). The critic network parameters \(\phi\) are trained by repeatedly sampling \(N\) times from the new buffer \(D\) to minimize the Bellman residual:
\begin{equation}
\label{15}
\begin{split}
% \mathcal{L}_{\phi} = \mathbb{E}^{\mathcal{D}} \left[ \left\| \left( R(a_k^0,s_k) + \gamma \hat{Q}_{\phi}(s_{t+1}, \epsilon_{\theta}(a_{k+1}^{t=0} | s_{k+1})) \right) \\
% - \hat{Q}_{\phi}(s_k, a_k^{t=0}) \right\|^2 \right]
\mathcal{L}_{critic}& = \mathbb{E}^{\mathcal{D}} \Bigg[ \Big\|  R(a_k^0,s_k) + \gamma_{\text{DIPO}} {Q}_{\phi}(s_{t+1}, \pi_{\theta}(a_{k+1}^{t=0} | s_{k+1}))    \\
&- {Q}_{\phi}(s_k, a_k^{t=0}) \Big\|^2 \Bigg].
\end{split}
\end{equation}

Then, the action gradient is performed for each action in the buffer \(D\), and the reward is regarded as a function of the action. All actions are updated along the direction of the action gradient using the gradient ascent method:
\begin{equation}
\label{16}
\begin{split}
a_k^{t=0} = a_k^{t=0} + \eta_{\text{DIPO}} \nabla_{a} {Q}_{\phi}(s_k, a_k^{t=0}).
\end{split}
\end{equation}

Then update the actor with the updated action:
\begin{equation}
\label{17}
\begin{split}
\mathcal{L}_{actor} = \mathbb{E}^{\mathcal{D}}\big[ \|\epsilon - \pi_{\theta}(a_k,s_k,t)\|^2\big].
\end{split}
\end{equation}

Diffusion Q-Learning (DQL) \cite{wang2022diffusion}: in DQL, the state-action Q function is learned and added to the training loss of the diffusion model to learn high-value actions:
\begin{equation}
\label{18}
\begin{split}
\mathcal{L}_{actor} = \mathbb{E}^{\mathcal{D}}\big[ \|\epsilon - \pi_{\theta}(a_k,s_k,t)\|^2 - \eta_{\text{DQL}}\hat{Q}_{\phi}(s_k, a_k)\big].
\end{split}
\end{equation}

Then, the Q-value function is minimized by using the double q-learning trick. In this process, DQL constructs two Q-networks \(Q_{\phi_{1}}\), \(Q_{\phi_{1}}\), and the target network \(\hat{Q}_{\phi_{1}}\), 
\(\hat{Q}_{\phi_{1}}\), and learns \({\phi}_1\), \({\phi}_1\) by minimizing the objective:
\begin{equation}
\label{19}
\begin{split}
\mathcal{L}_{critic}& = \mathbb{E}^{\mathcal{D}}\Bigg[ \Big\|  R(s_t, a_t^0) + \gamma \min_{i=1,2} \hat{Q}_{\phi_i}(s_{t+1}, a_{t+1}^0)  \\ &- {Q}_{\phi_i}(s_t, a_t^0)
\Big\|^2 \Bigg].
\end{split}
\end{equation}

Implicit Diffusion Q-learning (IDQL) \cite{hansen2023idql}: IDQL works similarly to DQL. DQL uses a diffusion model to parameterize actors and uses Q function to update actions. IDQL also updates policy by updating actions, but IDQL uses Q function and V function to reweight actions and finally forms the expected policy when resampling.

The goal of the value function is :
\begin{equation}
\label{20}
\begin{split}
\mathcal{L}_{v} = \mathbb{E}^{\mathcal{D}}\Bigg[ \Big\| \tau_{\text{IDQL}} - \mathfrak{E} \left( \hat{Q}_{\phi}(s_k, a_k^0) < {V}_{\psi}(s_k) \right) \Big\|^2 \Bigg].
\end{split}
\end{equation}

Then, this value function is used to update the Q function:
\begin{equation}
\label{21}
\begin{split}
\mathcal{L}_{q} = \mathbb{E}^{\mathcal{D}}\Bigg[ \Big\|  R(s_t, a_t) + \gamma {V}(s_k) - {Q}_(s_t, a_t)
\Big\|^2 \Bigg].
\end{split}
\end{equation}

In IDQL, it is proved that \(\pi_{\text{imp}}(a | s) \propto \pi_{\theta}(a | s) w(s, a)\), where the value of \(w(s, a)\) is determined by the \(V\) function and the \(Q\) function. There are three ways to calculate w(s, a) in \cite{hansen2023idql} (e.g., \(w_1^\tau(s,a) = \frac{|\tau - \mathbb{I}\left(Q(s,a) < V_1^\tau(s)\right)|}{|Q(s,a) - V_\tau^1(s)|}\)). Sample \(a_i\) from the learned diffusion policy \(\pi_{\theta}(a | s)\), calculate \(w(s, a_i)\) from the sampled samples, and finally update the policy to:
\begin{equation}
\label{22}
\begin{split}
p_i = \frac{w(s, a_i)}{\sum_j w(s, a_j)}.
\end{split}
\end{equation}

Q-score Matching (QSM) \cite{psenka2023learning}: QSM aims to establish a connection between the policy's score and the action gradient of the Q function, enabling the structure of the diffusion model policy to be aligned with the learned Q function. This linkage allows for optimizing the policy by matching its score with the Q function.

In QSM, the critic is updated using double Q learning, and its update formula is the same as Eq.~\eqref{19}.

The actor is updated to align with the gradient of the Q function:

\begin{equation}
\label{23}
\begin{split}
\mathcal{L}_{actor} = \mathbb{E}^{\mathcal{D}}\big[ \|\pi_{\theta}(a_k,s_k,t) - \alpha_{\text{QSM}} \nabla_a {Q}_{\phi} (s_k, a_k) \|^2\big].
\end{split}
\end{equation}

\subsection{Adam-based Diffusion Policy Optimization (ADPO) framework}
\label{ADPOFRAMEWORK}
Here, we present our ADPO framework in Algorithm \ref{alg:algorithm}.

\begin{algorithm}[H]
\caption{Adam-based Diffusion Policy Optimization Framework}\label{alg:algorithm}
\begin{algorithmic}
    \STATE {\textsc{Initialize}}: Critic network ${Q}_{\phi_1}$ (or ${Q}_{\phi_2}$, ${Q}_{\phi_3}$ for double Q learning) with random parameters $\phi_1$ (or $\phi_2$, $\phi_3$), target critic network parameters $\phi_1'$ (or $\phi_2'$, $\phi_3'$), replay buffer $D$, $\tau > 0$, pre-trained policy network $\pi_{\theta}$, episode, $num\_batch$.
    
    \FOR {each episode}
        \STATE $D \leftarrow \emptyset$
        \FOR {each $n\_step$}
            \STATE sample $a_k^0$ from $\pi_{\theta}(a_{k + 1} | s_{k+1})$ by Eq.~\eqref{4}
            \STATE step environment $r(s_{k + 1}| s_k, a_k^0)$, $s_{k + 1} \leftarrow$ \text{env}($a_t^0$)
            \STATE $D \leftarrow D \cup \{s_t, a_t^0, s_{t+1}, r(s_{t+1}|s_t, a_t^0)\}$
        \ENDFOR
        \STATE compute average episode reward, success rate
        \FOR {each $num\_batch$}
            \STATE sample $batch\_size$ from $D$
            \STATE compute advantage estimation for Eq.~\eqref{10}, \eqref{14}
            \STATE \textcolor{gray}{\# Compute Q-function learning loss}
            \STATE update critic ${Q}_{\phi_1}$ using Eq.~\eqref{12}, \eqref{15}, \eqref{20}, and \eqref{21} or ${Q}_{\phi_2}$ and ${Q}_{\phi_3}$ using Eq.~\eqref{19}
            \STATE \textcolor{gray}{\# Compute policy learning loss}
            \STATE update actor policy $\pi_{\theta}$ using Eq.~\eqref{10}, \eqref{11}, \eqref{14}, \eqref{17}, \eqref{18}, \eqref{22}, \eqref{23}
            \STATE \textcolor{gray}{\# Update Q-function and policy}
            \STATE $\theta_{0} = \theta$ or $\theta_{0} = {Q}_{\phi_1}$
            \FOR{$i = 1$ to $T$}
                \STATE $g_i \gets \nabla_{\theta} V^{\pi_{\theta}}(\rho) \Big|_{\theta = \theta_{i-1}}$
                \IF{using AdamW}
                    \STATE $\theta_i \gets \text{update via Eq.~\eqref{24}}$
                \ELSIF{using ADAPG}
                    \STATE $\theta_i \gets \text{update via Eq.~\eqref{25}}$
                \ENDIF
            \ENDFOR
        \ENDFOR
    \ENDFOR
\end{algorithmic}
\end{algorithm}

\noindent\textbf{Remark:} A few remarks of ADPO are provided here.
\begin{enumerate}
    \item We show that ADPO is a general diffusion-based RL framework. The ADAPG approach we use can accelerate a range of diffusion-based RL methods.
    \item The main difference between ADAPG and AdamW is that it takes into account the error accumulation caused by noise gradients. Inspired by Katyusha\cite{allen2018katyusha}, it uses the momentum parameter \(\omega\) to control the accuracy of iterative updates and accelerate the convergence of the algorithm.
\end{enumerate}

To more clearly highlight the advantages of our approach, we compare it with AdamW. 

AdamW updates the weight as follows:
\begin{equation}
\label{24}
\text{AdamW} : \begin{cases} 
m_i \gets \beta_1 m_{i-1} + (1 - \beta_1) g_i, \\ 
v_i \gets \beta_2 v_{i-1} + (1 - \beta_2) g_i^2, \\ 
\hat{m}_i \gets \frac{m_i}{1 - \beta_1^i}$, $\hat{v}_i \gets \frac{v_i}{1 - \beta_2^i}, \\
\theta_i \gets \theta_{i-1} - \eta \left( \frac{\hat{m}_i}{\sqrt{\hat{v}_i} + \varepsilon} + \lambda \theta_{i-1} \right),
\end{cases}
\end{equation}
where $\beta_1 \in (0, 1)$, $\beta_2 \in (0, 1)$, $\lambda > 0$, $\varepsilon > 0$, learning rate $\eta > 0$.

ADAPG updates the weight as follows:
\begin{equation}
\label{25}
\text{ADAPG} : \begin{cases} 
m_i \gets \beta_1 m_{i-1} + (1 - \beta_1) g_i, \\ 
v_i \gets \beta_2 v_{i-1} + (1 - \beta_2) g_i^2, \\ 
h_i \gets \theta_{i-1} - \eta \frac{(1 - \lambda) g_i + \lambda m_i}{\sqrt{v_i} + \varepsilon}, \\
\theta_i \gets \omega h_i + (1 - \omega) h_{i-1},
\end{cases}
\end{equation}
where $\omega \in (0,1.5]$.

\section{Experiment}
We apply ADPO into a series of diffusion-based RL methods (DPPO, DIPO, IDQL, DAWR, QSM, and DQL), coined ADPPO, ADIPO, AIDQL, ADAWR, AQSM, ADQL. We then conduct experiments with all 12 methods on the robotic benchmark tasks.  In addition, we perform numerical experiments on the hyperparameters 
$\varepsilon$ and $\omega$ for the ADPPO method in standard robot tasks to explore the numerical performance of these hyperparameters.

\subsection{Environments}
\label{Environments}
We confirm the efficacy of the resulting algorithmic framework by conducting experiments on three environments.

Environments I: We use three common benchmarks under OpenAI GYM \cite{brockman2016openai}: Hopper-v2 (controlling a bipedal robot to jump forward), Walker2D-v2 (controlling a bipedal robot to walk in a two-dimensional plane), and HalfCheetah-v2 (controlling a bipedal robot to run in a simulated environment).  
% The diffusion policy under the Gym task is pre-trained using the medium dataset in D4RL\cite{fu2020d4rl}, which is collected and generated by online training of Soft Actor-Critic\cite{haarnoja2018soft}. 

Environments II: ROBOMIMIC\cite{mandlekar2021matters}. 
% We use the Multi-Human (MH) dataset with 300 noisy human demonstrations and 100 demonstrations provided by ROBOMIMIC for pre-training, from which 
We select three robot manipulation tasks: Lift (lifting a cube from the table), Can (picking up a Coke can and placing it at a target bin), Square (picking up a square nut and place it on a rod), with difficulty ranging from low to high.

\begin{figure}[!t]
    \centering
    \subfloat[]{%
        \includegraphics[width=0.15\textwidth]{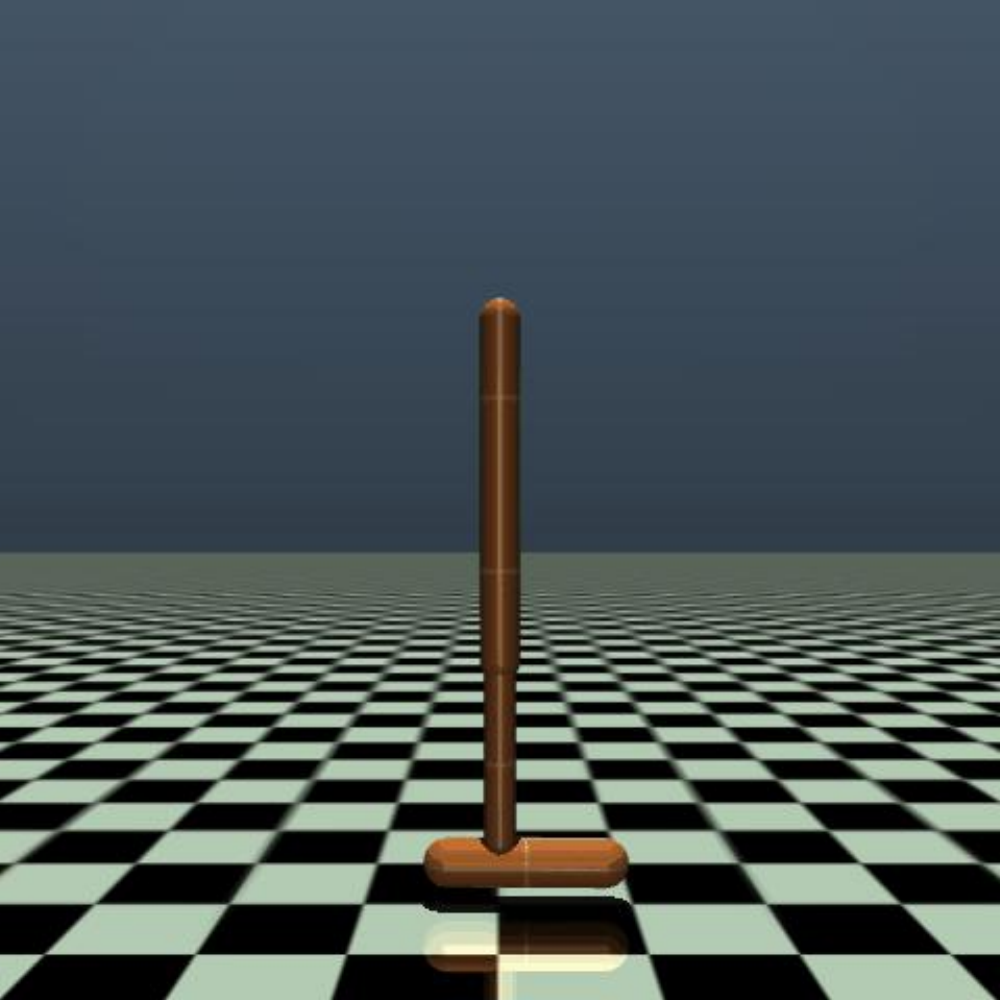}
        \label{fig_1a}
    }
    \hfill
    \subfloat[]{%
        \includegraphics[width=0.15\textwidth]{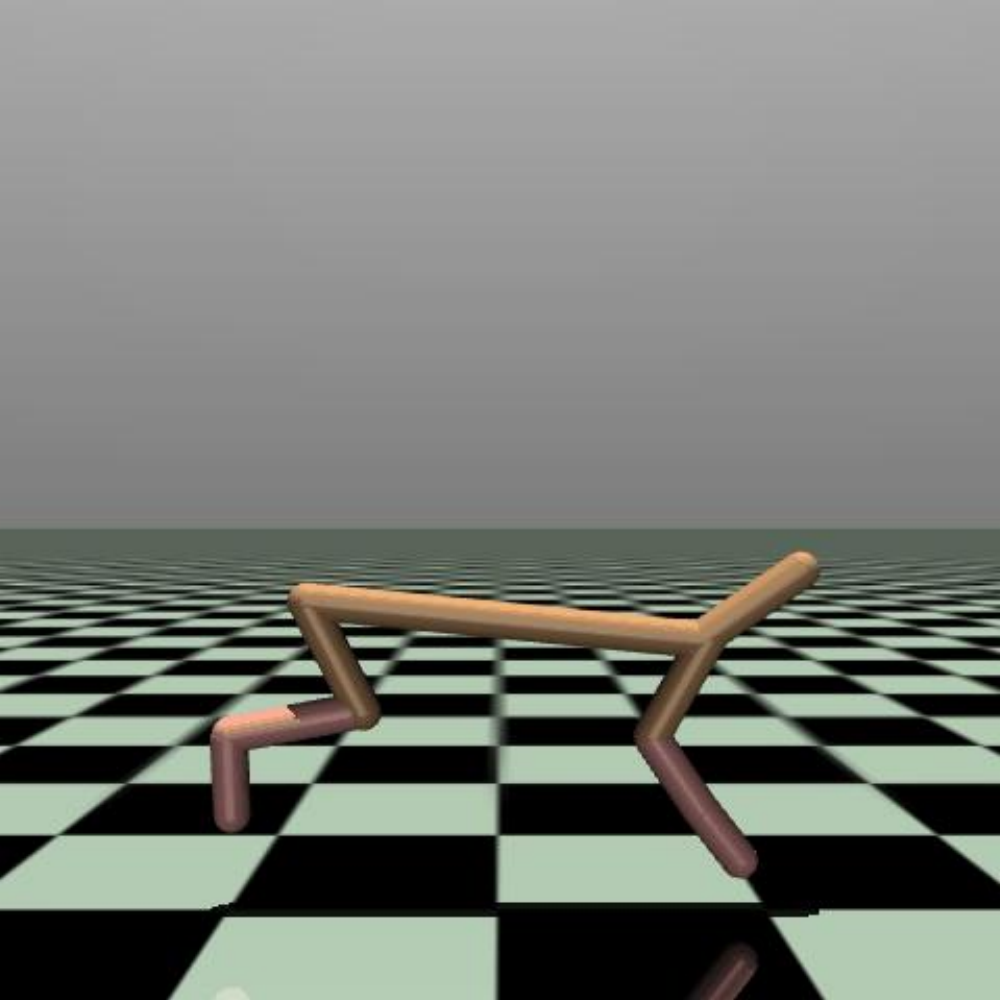}
        \label{fig_1b}
    }
    \label{fig_1}
    \subfloat[]{%
        \includegraphics[width=0.15\textwidth]{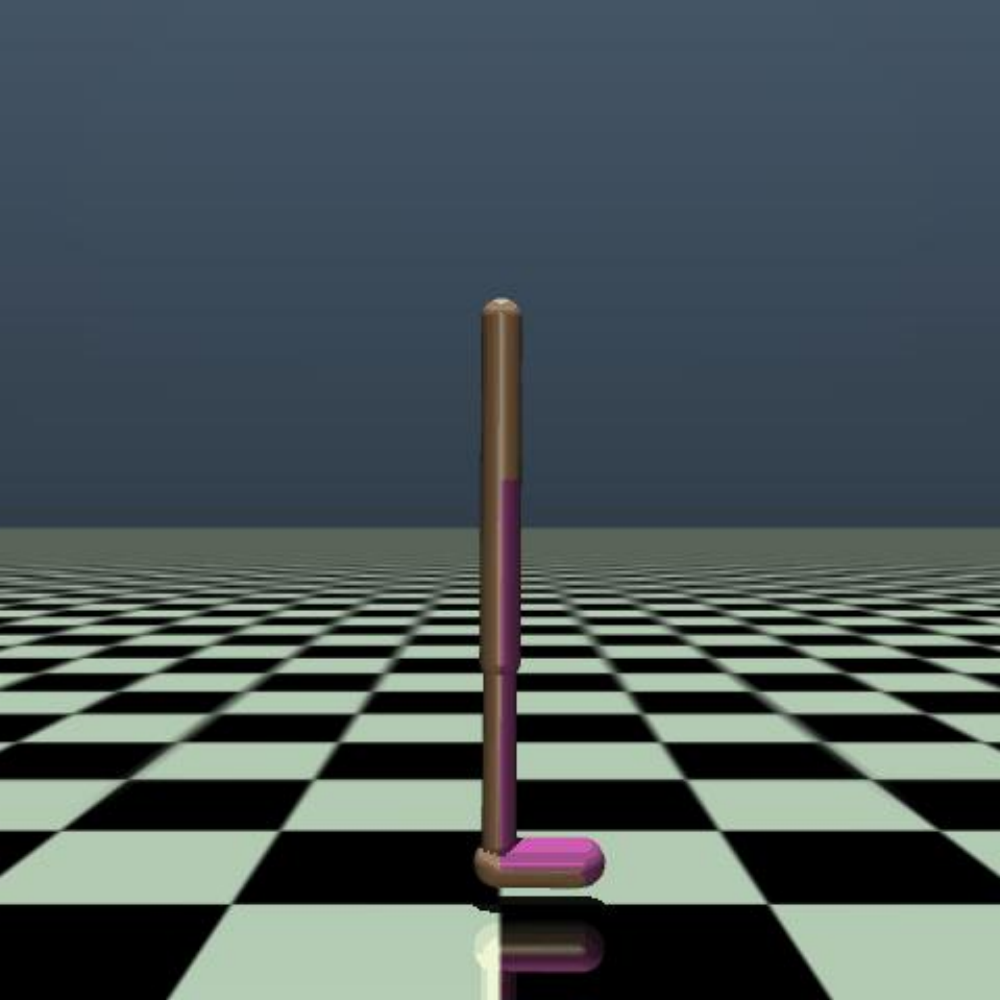}
        \label{fig_1c}
    }
    \hfill
    \subfloat[]{%
        \includegraphics[width=0.15\textwidth]{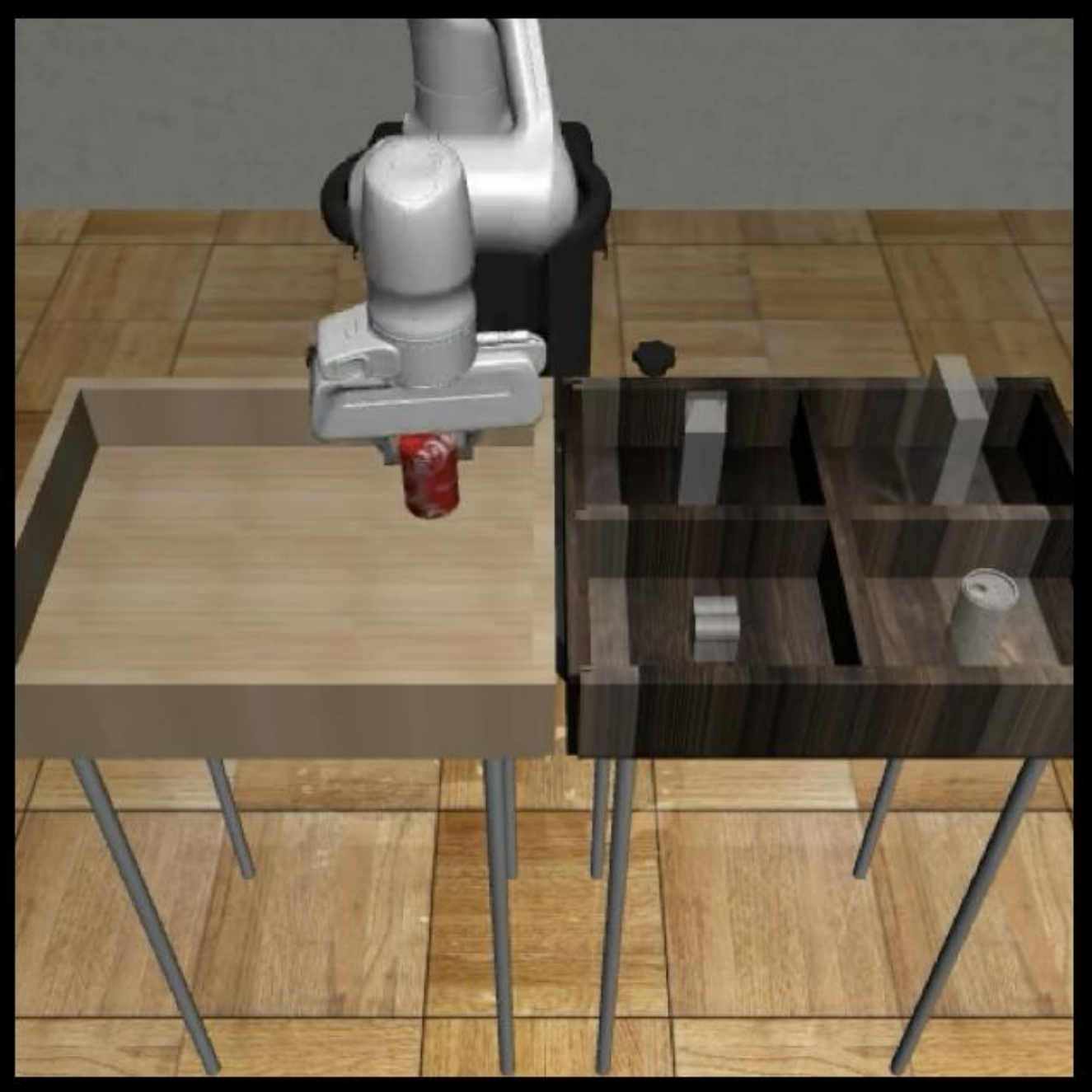}
        \label{fig_1d}
    }
    \label{fig_2}
    \subfloat[]{%
        \includegraphics[width=0.15\textwidth]{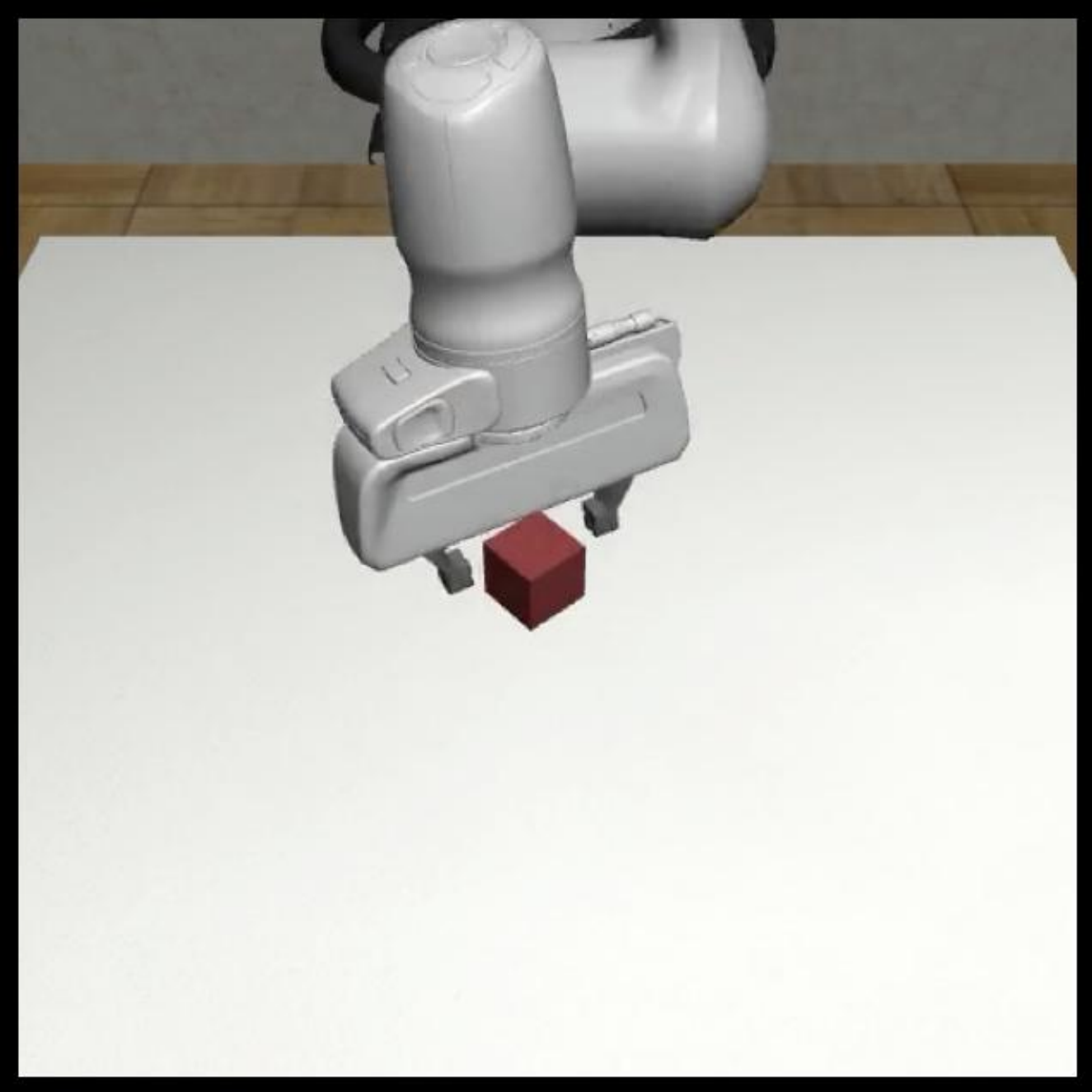}
        \label{fig_1e}
    }
    \hfill
    \subfloat[]{%
        \includegraphics[width=0.15\textwidth]{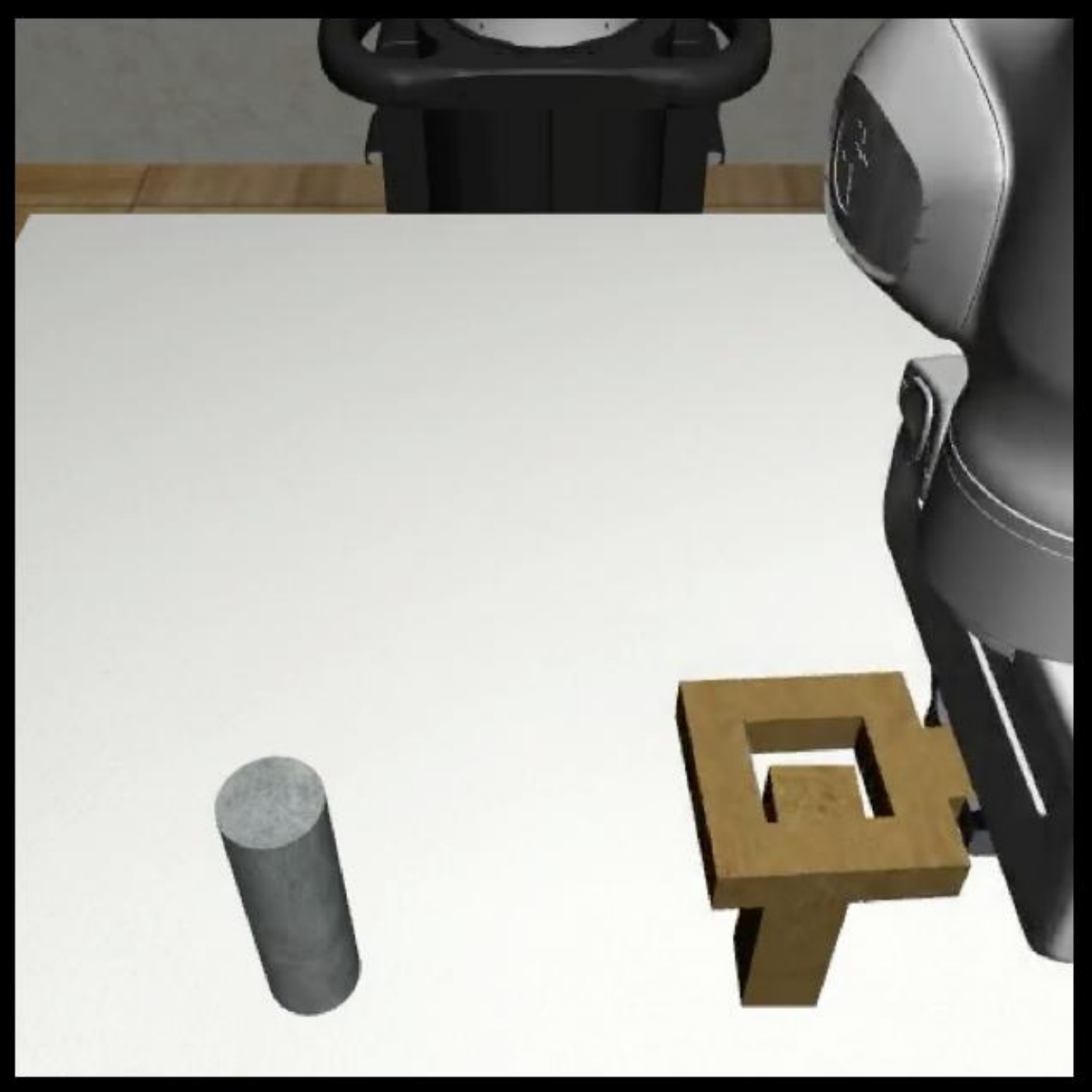}
        \label{fig_1f}
    }
    \label{fig_3}
    \caption{The first line shows the case of GYM benchmark, and the second line shows the case of ROBOMIMIC benchmark.}
    \label{fig_4}
\end{figure}

\textbf{Experimental details.} For pre-training dataset, all observations and actions are normalized to the range $[-1, 1]$. For GYM tasks, pre-training policies are trained for 3000 iterations with a batch size of 128, a learning rate of $1 \times 10^{-3}$, which is decayed to $1 \times 10^{-4}$ using cosine scheduling, and a weight decay of $1 \times 10^{-6}$. For ROBOMIMIC tasks, pre-training policies are trained for 8000 iterations with a batch size of 128, a learning rate of $1 \times 10^{-4}$, which is decayed to $1 \times 10^{-5}$ using cosine scheduling, and a weight decay of $1 \times 10^{-6}$. 
For formal fine-tuning, the parameter settings in each task are different. The detailed configuration is shown in Table \ref{appendix-1} in the Appendix.

\subsection {Comparison of ADPO diffusion-based RL methods with Baselines}
\label{ADAPG on diffusion-based RL algorithms}
We experiment with 12 methods on three GYM tasks and three ROBOMIMIC tasks. In the ADPO diffusion-based RL methods, the value of $\varepsilon$ = $10^{-11}$, the value of \(\omega \in (0, 1.5]\), and other hyperparameters are the same as the baselines. 

In Figures 2 and 3, to facilitate the comparison between the ADAPG diffusion-based RL methods and the diffusion-based RL methods, we use two figures to show the operation of these 12 methods on the same dataset (i.e., Figures 1.a and 1.b are the results under Hopper-v2, Figures 1.c and 1.d are the results under Halfcheetah-v2 and Figures 1.e and 1.f are the results under Walker2d-v2). The shadows in Figures 2 and 3 represent the standard deviation across 5 and 3 seeds, respectively. The horizontal axis represents the time step, and the vertical axis represents the average reward or success rate obtained by the agent when interacting with the environment. The dotted line is our ADPO diffusion-based RL methods, and the solid line represents a series of diffusion-based RL methods. For ease of comparison, the same color is used. The vertical axis indicators of GYM tasks and ROBOMIMIC tasks are different. Because GYM tasks are discrete actions and average reward can better reflect the quality of the robot's actions, while ROBOMIMIC tasks are continuous actions, success rate can better reflect the quality of the robot's actions.

In general, ADPO can significantly improve the performance of a series of diffusion-based RL methods, especially when facing complex tasks, showing good stability and efficient policy learning ability. In GYM tasks, ADPPO, ADIPO and ADQL all show more competitive performance than the original baselines. In more challenging ROBOMIMIC tasks, ADPO can still maintain the stability of training, improve the convergence speed of the algorithm in continuous action space tasks, and can effectively handle long-distance tasks.

\begin{table}[h]  % 使用table环境，表示在单栏内
    \centering
    \caption{The comparison results between ADPO and baseline optimizers on GYM tasks and ROBOMIMIC tasks. Y means above the baselines, and \(-\) is on par with the baselines.}
    \resizebox{\columnwidth}{!}{  % 将表格宽度调整为单栏的宽度
    \begin{tabular}{lcccccc}
        \toprule
        Task & ADPPO & ADIPO & AIDQL & ADQL & AQSM & ADAWR  \\
        \midrule
        hopper       & Y & Y & Y & Y & -- & Y   \\
        halfcheetah  & Y & Y & -- & Y & -- & --  \\
        walker2d     & Y & Y & -- & Y & Y & --   \\
        can          & Y  & Y  & -- & -- & Y & --    \\
        lift         & -- & -- & Y & -- & Y & Y  \\
        square       & Y & Y & Y & -- & -- & Y  \\
        \bottomrule
    \end{tabular}
    }
    \label{tab:comparison}
\end{table}

\begin{figure}[!t]
    \centering
    \subfloat[]{%
        \includegraphics[width=0.23\textwidth]{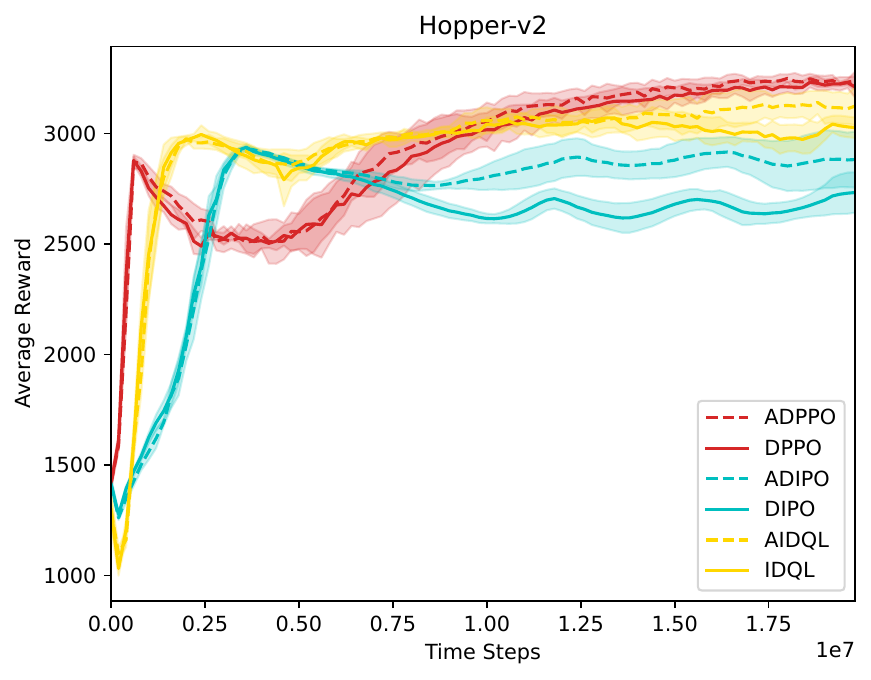}
        \label{fig_2a}
    }
    \hfill
    \subfloat[]{%
        \includegraphics[width=0.23\textwidth]{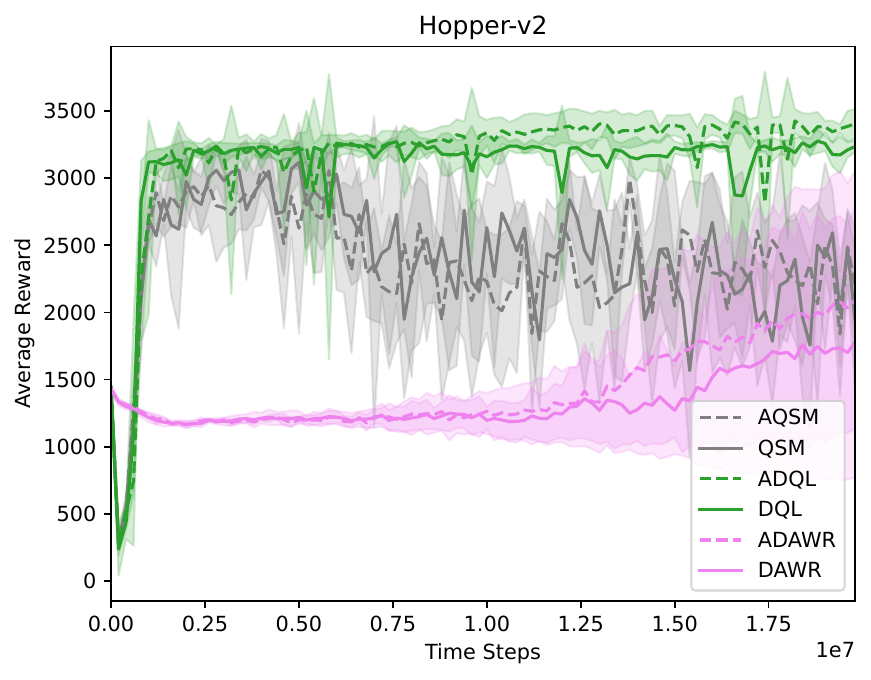}
        \label{fig_2b}
    }
    \label{fig_5}
    \subfloat[]{%
        \includegraphics[width=0.23\textwidth]{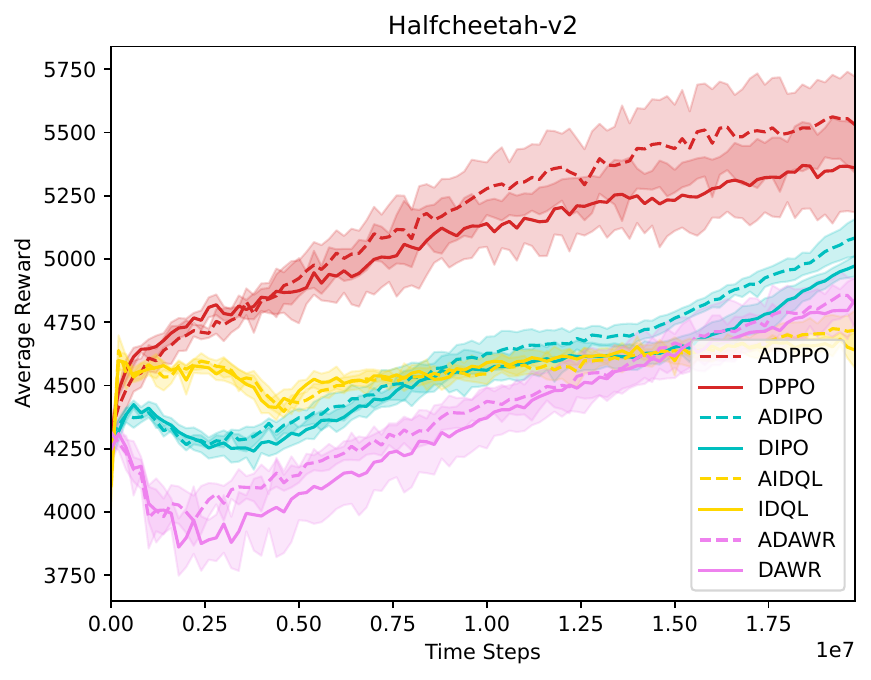}
        \label{fig_2c}
    }
    \hfill
    \subfloat[]{%
        \includegraphics[width=0.23\textwidth]{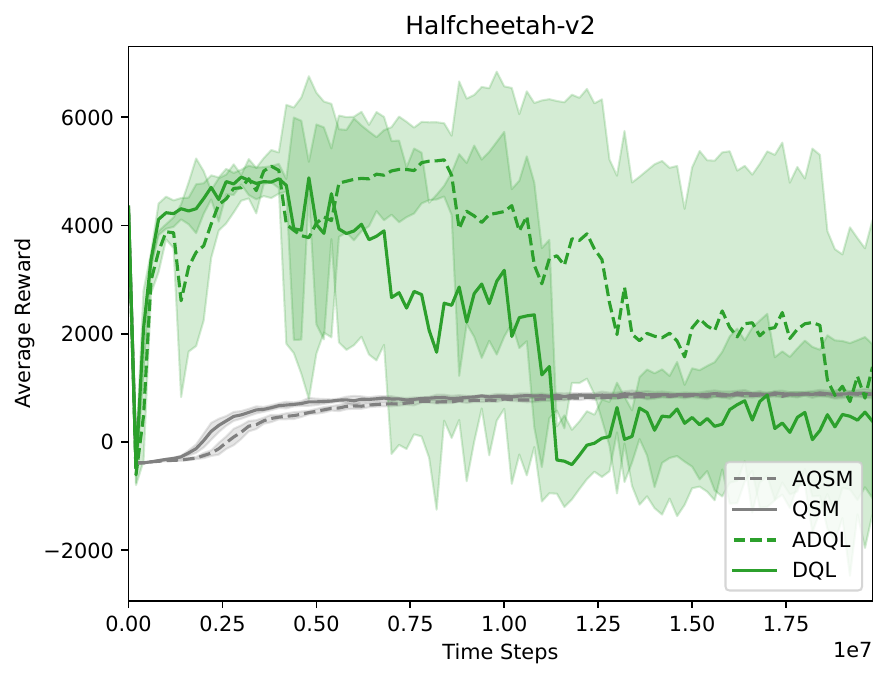}
        \label{fig_2d}
    }
    \label{fig_6}
    \subfloat[]{%
        \includegraphics[width=0.23\textwidth]{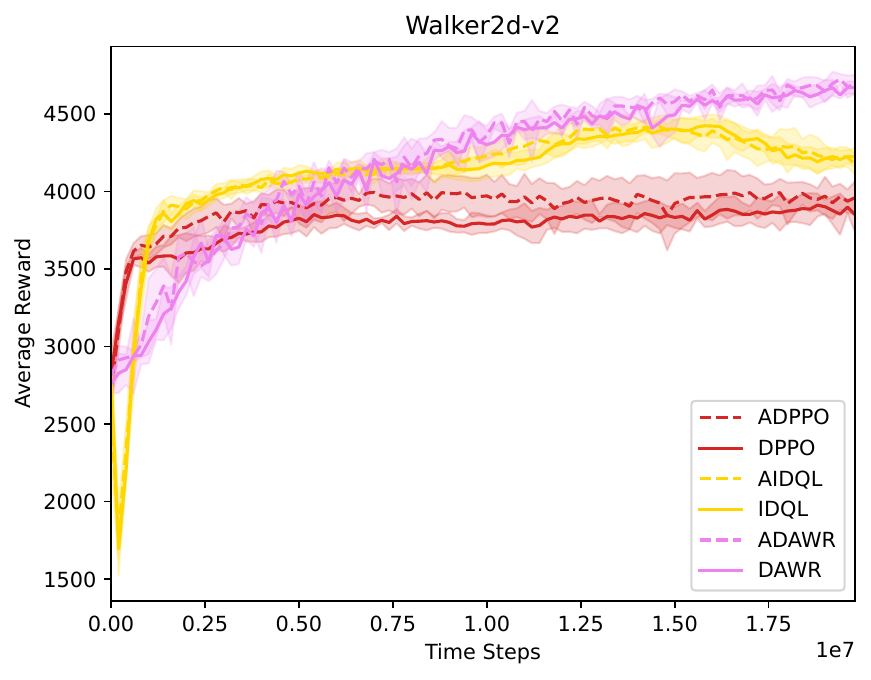}
        \label{fig_2e}
    }
    \hfill
    \subfloat[]{%
        \includegraphics[width=0.23\textwidth]{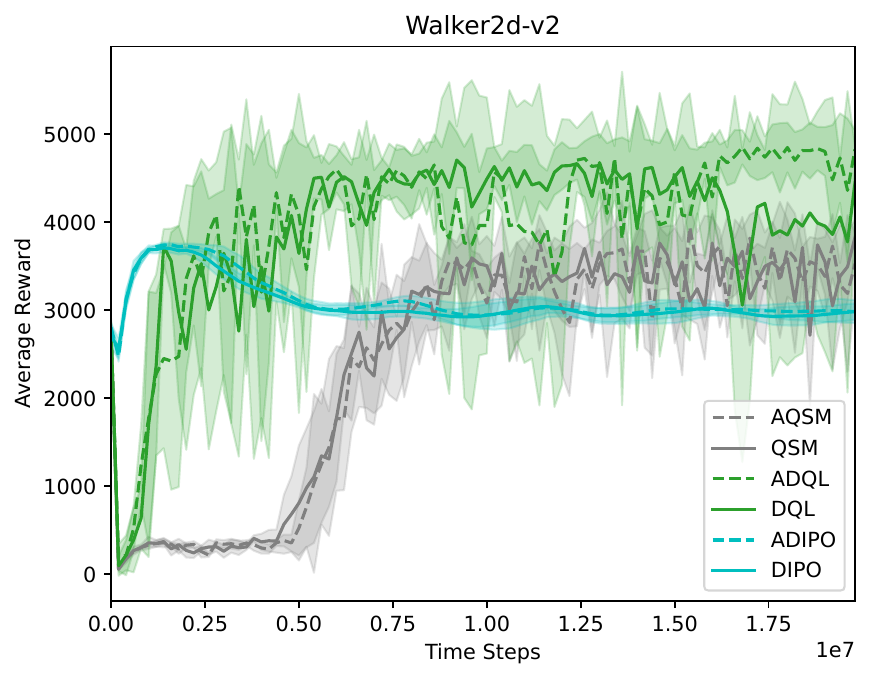}
        \label{fig_2f}
    }
    \label{fig_7}
    \caption{Comparison of ADAPG diffusion-based RL methods with modern diffusion-based RL methods on the GYM tasks, with an average of 5 seeds.}
    \label{fig_8}
\end{figure}

\begin{figure}[!t]
    \centering
    \subfloat[]{%
        \includegraphics[width=0.23\textwidth]{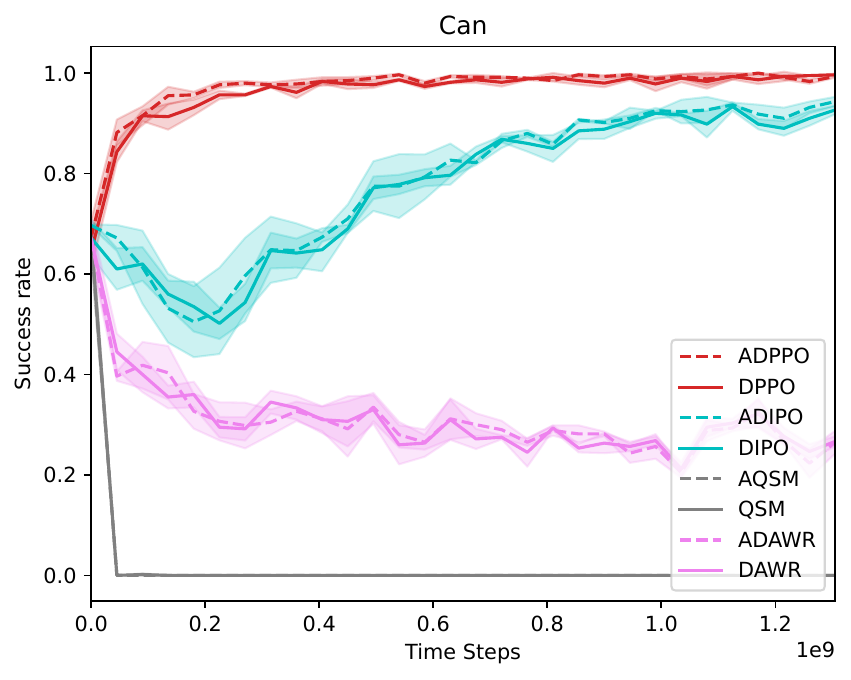}
        \label{fig_3a}
    }
    \hfill
    \subfloat[]{%
        \includegraphics[width=0.23\textwidth]{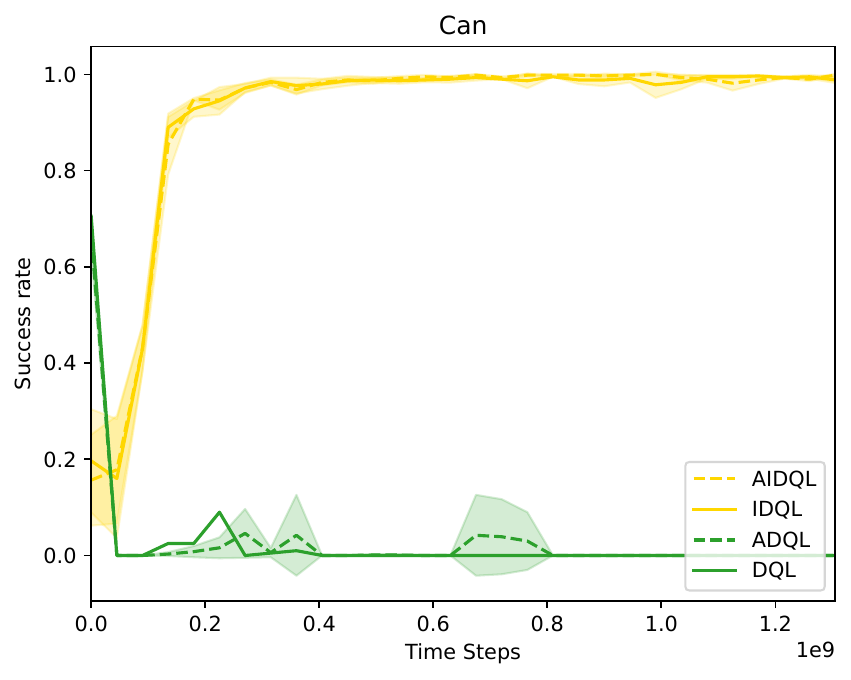}
        \label{fig_3b}
    }
    \label{fig_9}
    \subfloat[]{%
        \includegraphics[width=0.23\textwidth]{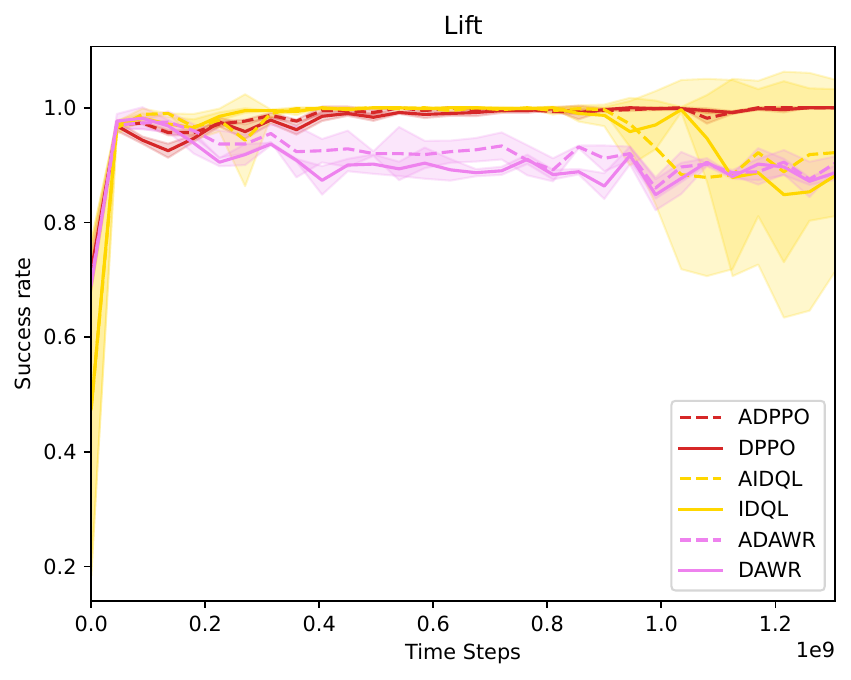}
        \label{fig_3c}
    }
    \hfill
    \subfloat[]{%
        \includegraphics[width=0.23\textwidth]{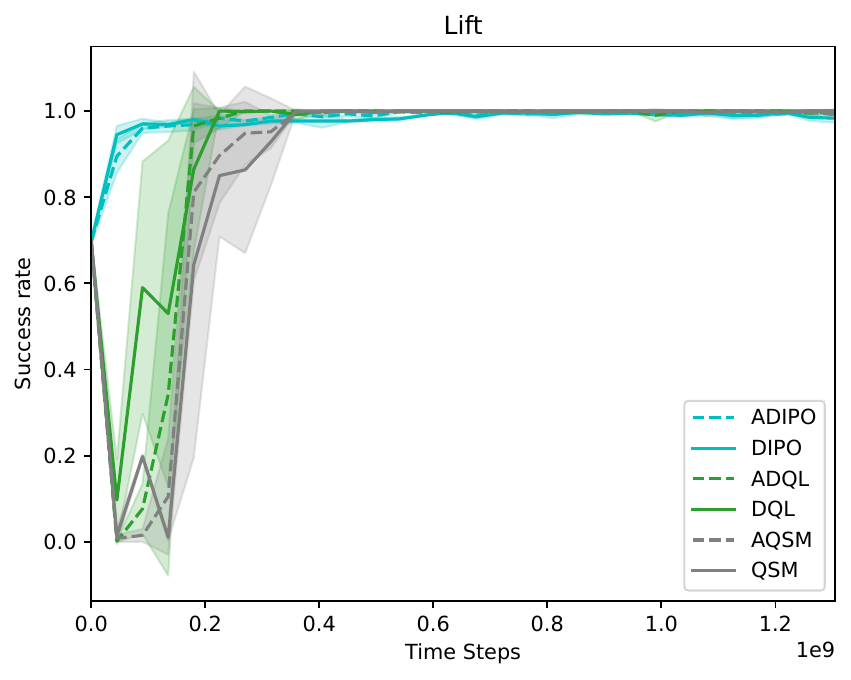}
        \label{fig_3d}
    }
    \label{fig_10}
    \subfloat[]{%
        \includegraphics[width=0.23\textwidth]{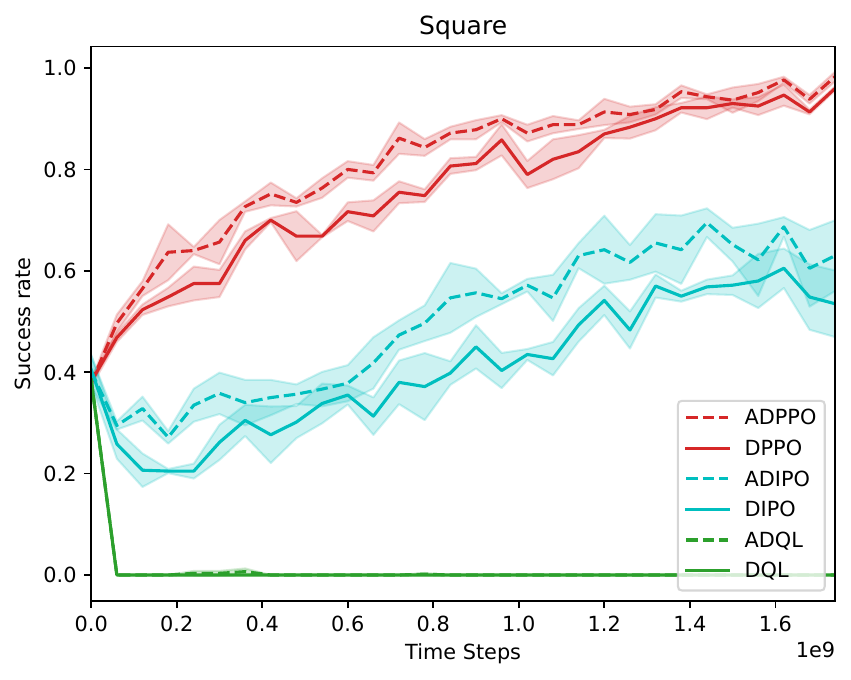}
        \label{fig_3e}
    }
    \hfill
    \subfloat[]{%
        \includegraphics[width=0.23\textwidth]{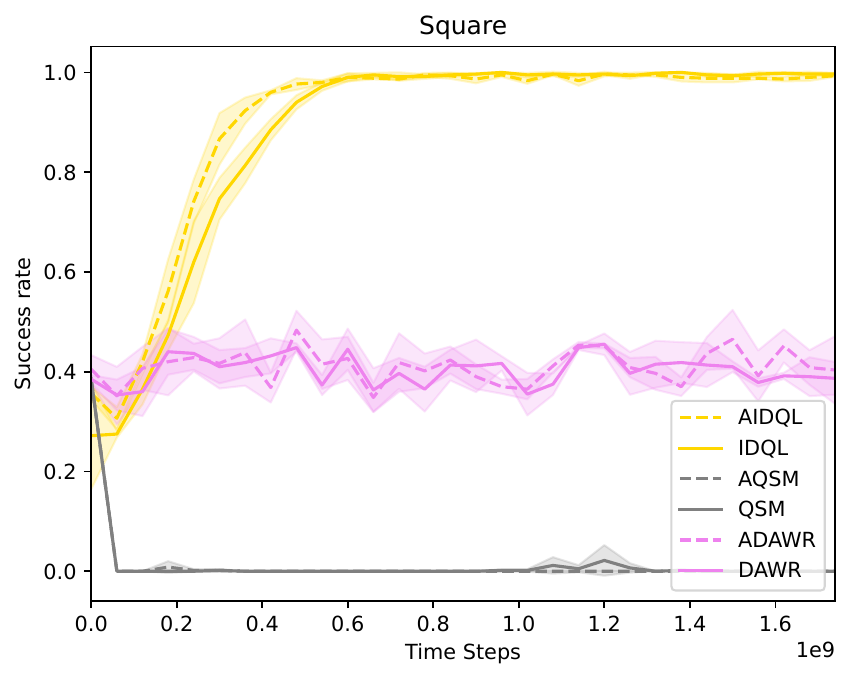}
        \label{fig_3f}
    }
    \label{fig_11}
    \caption{Comparison of ADAPG diffusion-based RL methods with advanced diffusion-based RL methods on the ROBOMIMIC tasks, with an average of 3 seeds.}
    \label{fig_12}
\end{figure}

\subsection{Performance of different $\varepsilon$ and $\omega$ on GYM tasks}
\label{Performance of different on Halfcheetah-v2}
To investigate the impact of $\varepsilon$ and $\omega$ on ADPO, we design ablation experiments to determine the optimal choices for two hyperparameters. The ablation results are shown in Figure 3. We find that the hyperparameter $\varepsilon$ achieves consistently strong performance across multiple tasks when set to $10^{-11}$. This indicates that $\varepsilon$ is robust and doesn't
require extensive tuning for different environments. This indicates that $\varepsilon$ is robust and doesn't require extensive tuning for different environments. The hyperparameter $\omega$ is sensitive to the environment, with its optimal value varying across different environments and methods. The values of $\omega$ for different tasks are shown in Table \ref{tab:comparison}.

\begin{figure}[!t]
    \centering
    \subfloat[]{%
        \includegraphics[width=0.23\textwidth]{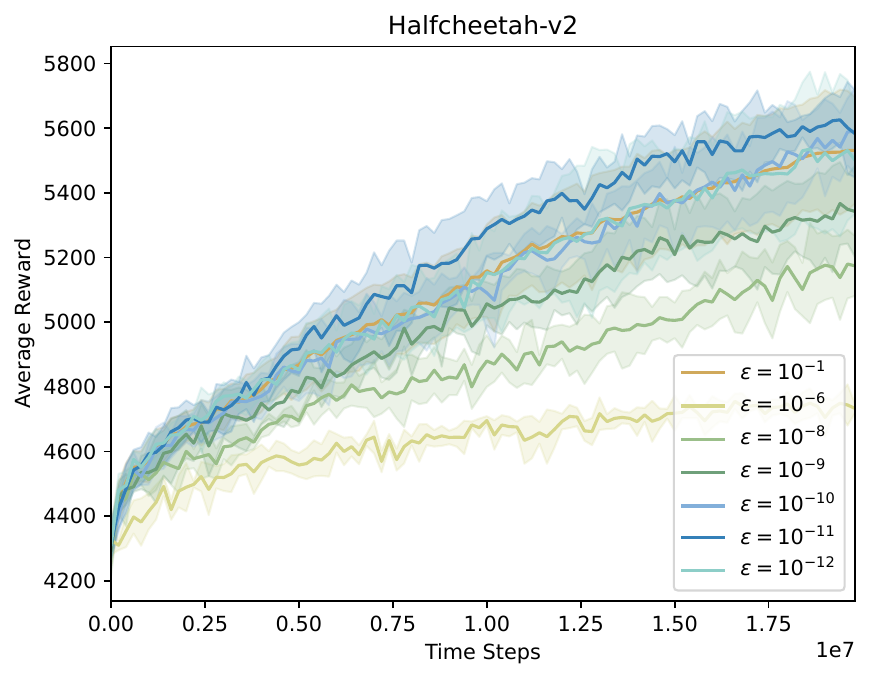}
        \label{fig_4a}
    }
    \hfill
    \subfloat[]{%
        \includegraphics[width=0.23\textwidth]{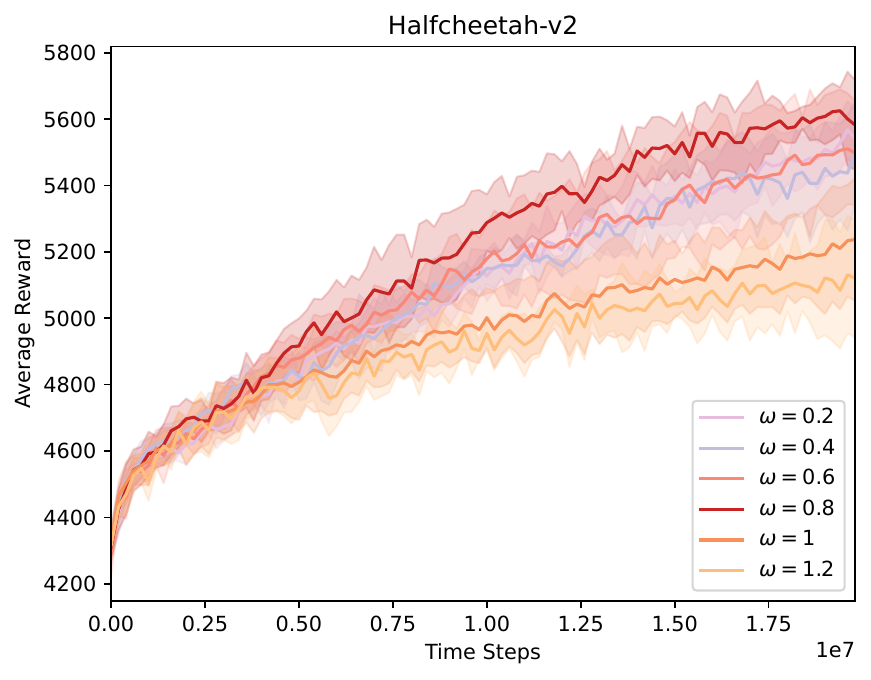}
        \label{fig_4b}
    }
    \label{fig_13}
    \subfloat[]{%
        \includegraphics[width=0.23\textwidth]{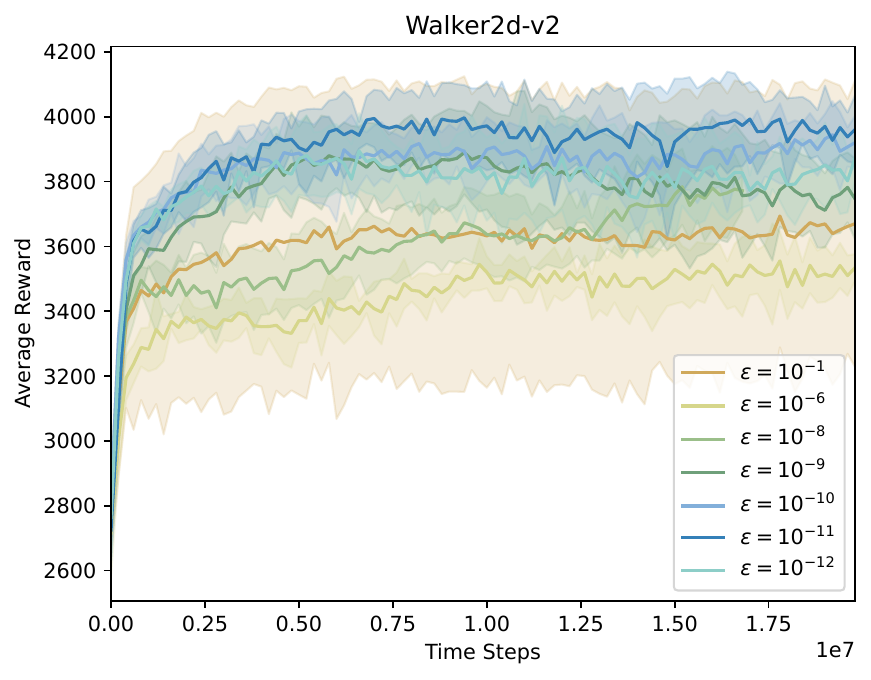}
        \label{fig_4c}
    }
    \hfill
    \subfloat[]{%
        \includegraphics[width=0.23\textwidth]{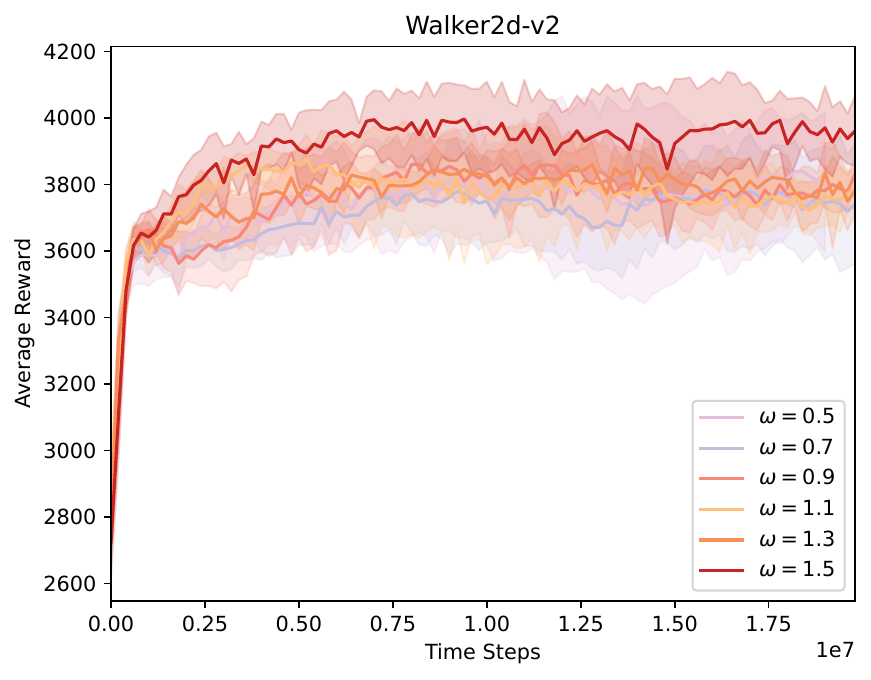}
        \label{fig_4d}
    }
    \label{fig_14}
    \caption{A comparison of the performance of ADPPO with different $\varepsilon$ (left) and different $\omega$ (right) on GYM tasks.}
    \label{fig_15}
\end{figure}

\begin{table}[h]  % 使用table环境，表示在单栏内
    \centering
    \caption{The setting of $\omega$ on diffusion-based RL methods under different tasks.}
    \resizebox{\columnwidth}{!}{  % 将表格宽度调整为单栏的宽度
    \begin{tabular}{lcccccc}
        \toprule
        Task & ADPPO & ADIPO & AIDQL & ADQL & AQSM & ADAWR  \\
        \midrule
        hopper       & 1.2 & 1.5 & 0.8 & 1 & 1.2 & 1.2 \\
        halfcheetah  & 0.8 & 1.2 & 0.8 & 0.6 & 0.8 & 1.2 \\
        walker2d     & 1.5 & 1.5 & 1.2 & 1 & 1 & 1.5 \\
        can          & 1.5  & 1  & 1 & 1 & 1  & 1 \\
        lift         & 1.5 & 1.5 & 1.5 & 1.5 & 1.2  & 1.5 \\
        square       & 1.5 & 1.5 & 1.5 & 1.5 & 1.5 & 1.2 \\
        \bottomrule
    \end{tabular}
    }
    \label{tab:comparison}
\end{table}

\section{CONCLUSION}
To solve the challenge of how to optimize diffusion-based polices fast and stably, this work considered the use of adaptive gradient methods in RL, leading to a fast algorithmic framework containing best practices for fine-tuning diffusion-based polices in robotic control tasks, coined ADPO. To verify the effectiveness of ADPO, we compared it with six advanced diffusion-based RL methods and conducted extensive experiments in various robotics benchmarks. The results showed that ADPO not only accelerated the training process, but also achieved better performance than other baseline methods on multiple tasks, especially in cases of high task complexity or large environmental changes. More importantly, we systematically analyzed the sensitivity of multiple hyperparameters in standard robotics tasks, providing guidance for subsequent practical applications.

\bibliographystyle{IEEEtran}
\bibliography{ADPO}

% % \newpage

{\appendix[Settings of the Experiment]

For the parameter settings of different tasks, including GYM and ROBOMIMIC, we referred readers to Table \ref{appendix-1}.

\begin{table*}[b]
\caption{Hyper-parameter settings of GYM and ROBOMIMIC}
\label{appendix-1}
\centering
\begin{tabular}{lcccccccc}
\toprule
Environment & Task & Number of envs & Obs dim & Act dim & \( T \) & Actor learning rate & Critic learning rate & Batch size \\
\midrule
GYM & Hopper-v2 & 40 & 11  & 3 & 1000 & \( 10^{-4} \) & \( 10^{-3} \) & 1000 \\
 & Walker2D-v2 & 40 & 17 & 6 & 1000 & \( 10^{-4} \) & \( 10^{-3} \) & 1000 \\
 & Halfcheetah-v2 & 40 & 17  & 6 & 1000 & \( 10^{-4} \) & \( 10^{-3} \) & 1000 \\
\midrule
ROBOMIMIC & Lift & 50 & 19 & 7 & 300 & \( 10^{-5} \) & \( 10^{-3} \) & 1000 \\
& Can & 50 & 23 & 7 & 300 & \( 10^{-5} \) & \( 10^{-3} \) & 1000 \\
& Square & 50 & 23 & 7 & 300 & \( 10^{-5} \) & \( 10^{-3} \) & 1000 \\
\midrule
\end{tabular}
\label{tab:task_dataset}
\end{table*}
}
\vfill

\end{document}